%% file: icassp23-main.tex
\documentclass{article}
\usepackage{spconf}

\usepackage{times}

\usepackage{soul}
\usepackage{url}
\usepackage[hidelinks]{hyperref}
\usepackage[utf8]{inputenc}
\usepackage[small]{caption}
\usepackage{graphicx}
\usepackage{amsmath}
\usepackage{amsthm}
\usepackage{booktabs}
\usepackage{multirow}

\usepackage{amsfonts,amssymb}
\DeclareSymbolFontAlphabet{\mathbb}{AMSb}
\usepackage{float}    
\usepackage{fancyhdr}
\usepackage{fancybox}
\usepackage{xspace}
\usepackage{array}
\usepackage{multirow}
\usepackage[mathscr]{euscript}
\usepackage{subcaption}
\usepackage{bookmark}

\usepackage{xcolor}
\usepackage{color}
\usepackage{pgfplots}
\usepackage{pgfplotstable}
\usepackage{hhline}
\usepackage{epstopdf}
\usepackage{stmaryrd}

\usepackage[T1]{fontenc}    
\usepackage{hyperref}       

\usepackage{booktabs}       

\usepackage{nicefrac}       
\usepackage{microtype}      

\usepackage{graphicx}

\newcommand{\vertiii}[1]{{\left\vert\kern-0.25ex\left\vert\kern-0.25ex\left\vert #1 
		\right\vert\kern-0.25ex\right\vert\kern-0.25ex\right\vert}}
\input{mymacros_2cln}
\input{lpsymbols}

\newcommand{\lb}{\llbracket}
\newcommand{\rb}{\rrbracket}

\title{Distributionally Robust Multiclass Classification and Applications in Deep Image Classifiers}

\name{Ruidi Chen$^{\star}$ \qquad Boran Hao$^{\dagger}$ \qquad Ioannis Ch. Paschalidis$^{\dagger}$}

\address{$^{\star}$Amazon SCOT \\
$^{\dagger}$Department of Electrical and Computer Engineering, Boston University}

\begin{document}
\maketitle
\begin{abstract}
    We develop a {\em Distributionally Robust Optimization (DRO)} formulation for {\em Multiclass Logistic Regression (MLR)}, which could tolerate data contaminated by outliers. The DRO framework uses a probabilistic ambiguity set defined as a ball of distributions that are close to the empirical distribution of the training set in the sense of the Wasserstein metric. We relax the DRO formulation into a regularized learning problem whose regularizer is a norm of the coefficient matrix. We establish out-of-sample performance guarantees for the solutions to our model, offering insights on the role of the regularizer in controlling the prediction error. We apply the proposed method in rendering deep {\em Vision Transformer (ViT)}-based \cite{vit} 
    image classifiers robust to random and adversarial attacks. Specifically, using the MNIST and CIFAR-10 datasets, we demonstrate reductions in test error rate by up to 83.5\% and loss by up to 91.3\% compared with baseline methods, by adopting a novel random training method. 
    
\end{abstract}

\section{Introduction} \label{sec:intro}
    We consider the robust multi-class classification problem under the framework of {\em
	Distributionally Robust Optimization (DRO)}, where the ambiguity set is defined via
the Wasserstein metric \cite{gao2016distributionally,esfahani2018data}.
Robust optimization has been widely used for inducing robustness to classification models \cite{el2003robust,bertsimas2018robust}. 
DRO, which minimizes the worst-case loss over a probabilistic ambiguity set, has received an increasing attention in recent years.
The ambiguity set in DRO can be defined through moment constraints~\cite{Sim14}, or as a ball of distributions using some probabilistic distance function such as the Wasserstein distance \cite{esfahani2018data}.
The Wasserstein DRO model has been extensively studied in the machine learning community \cite{chen2018robust,blanchet2019multivariate,sinha2017certifiable,abadeh2015distributionally,shafieezadeh2017regularization,gao2017wasserstein,OPT-026}.

Most of the works on distributionally robust classification have focused on the binary setting \cite{abadeh2015distributionally}. In this paper, we extend the DRO framework to the multi-class setting by exploring {\em Multiclass Logistic Regression (MLR)} and deriving the corresponding robust model.
We obtain a novel {\em matrix norm} regularizer for MLR through reformulating the DRO problem; thus, establishing a connection between robustness and regularization, and enabling a primal-dual interpretation for the data-regularizer relationship. Note that the link between robustness and regularization has been established in the 1-dimensional output setting, see, e.g., \cite{LAU97,el2003robust,xu2009robustness,xu2009robust,bertsimas2017characterization} for deterministic disturbances, and \cite{abadeh2015distributionally,chen2018robust,blanchet2019multivariate,shafieezadeh2017regularization,gao2017wasserstein} for disturbances within a Wasserstein set. However, none of these works studied the robust problem with multi-dimensional outputs.

To the best of our knowledge, we are the first to study the robust multi-class classification problem from the standpoint of Wasserstein distributional robustness. Our problem is closely related to \cite{abadeh2015distributionally,chen2018robust} where the Wasserstein DRO problem with a 1-dimensional output was studied. The key differences lie in that: 
(i) \cite{abadeh2015distributionally,chen2018robust} only considered
a scalar output $y$. We make the non-trivial extension to a
multi-class scenario, which is very different and our key and
novel contribution. Specifically, in this work we are regularizing
a coefficient matrix $\bB$ and the correlation structure
embedded in the responses should be reflected in the regularizer
which is derived as the dual norm of the matrix; (ii) the out-of-sample result in Theorem \ref{c1} is tailored to the classification setting, which is different from the regression setting in \cite{chen2018robust}, and from 
\cite{abadeh2015distributionally} which bounds the out-of-sample risk by the optimal worst-case risk; and (iii) we demonstrate a computationally efficient random-layer training mechanism for applying DRO in {\em Vision Transformer (ViT)}-based \cite{vit} image classifiers, which adds to the accessibility and appeal of this work to practitioners.

%

The rest of the paper is organized as follows. In Section~\ref{s2}, we
develop the Wasserstein DRO formulation for MLR. Section~\ref{s3}
establishes the out-of-sample performance guarantees for the DRO solution. Numerical experimental results with deep {\em Vision Transformer (ViT)}-based  \cite{vit} 
image classifiers are presented in
Section~\ref{s4}. We conclude the paper in Section~\ref{s5}.

\textbf{Notational convention.} We use boldfaced lowercase letters to
denote vectors, ordinary lowercase letters to denote scalars, boldfaced
uppercase letters to denote matrices, and calligraphic capital letters
to denote sets. 
All vectors are column vectors. 
For space
saving reasons, we write $\bx=(x_1, \ldots, x_{\text{dim}(\bx)})$ to
denote the column vector $\bx$, where $\text{dim}(\bx)$ is the dimension
of $\bx$. We use prime to denote the transpose, $\|\cdot\|_p$
for the $\ell_p$ norm with $p \ge 1$, 
and $\|\bx\|_p^{\bW}$ for the $\bW$-weighted $\ell_p$ norm defined as $\|\bx\|_p^{\bW} \triangleq \|\bW^{1/2}\bx\|_p$, with a positive definite matrix $\bW$. For a matrix $\bA \in \mbb{R}^{m \times n}$, we use $\|\bA\|_p$ to denote its induced $\ell_p$ norm that is defined as $\|\bA\|_p \triangleq \sup_{\bx \neq \mathbf{0}}\|\bA\bx\|_p/\|\bx\|_p$. 
Finally, $\mathbf{1}$ denotes the vector of ones, $\mathbf{0}$ the vector of zeros,
and $\mathbf{e}_k$ the $k$-th unit vector.

\section{Problem Formulation} \label{s2}
Suppose there are $K$ classes, and we are given a predictor $\bx \in \mbb{R}^p$. Our goal is to predict its class label, denoted by a $K$-dim vector $\by \in \{0, 1\}^K$, where $\by = (y_1, \ldots, y_K)$, $\sum_k y_k = 1$, and $y_k=1$ if and only if $\bx$ belongs to class $k$, in which case $\by = \mathbf{e}_k$. The conditional distribution of $\by$ given $\bx$ is modeled as
$p(\by|\bx) = \prod_{i=1}^K p_i^{y_i},$
where $p_i = e^{\bw_i'\bx}/\sum_{k=1}^K e^{\bw_k'\bx}$, and $\bw_i, i \in \lb K \rb$, are the coefficient vectors to be estimated.  The log-likelihood can be expressed as:
\begin{equation*}
\log p(\by|\bx)  = \sum_{i=1}^K y_i \log(p_i) 
= \by'\bB'\bx - \log \mathbf{1}'e^{\bB'\bx},
\end{equation*} 
where $\bB \triangleq [\bw_1 \cdots \bw_K]$ and the exponential operator is applied element-wise. The log-loss is defined as $h_{\bB}(\bx, \by) \triangleq \log \mathbf{1}'e^{\bB'\bx} - \by'\bB'\bx$. The Wasserstein DRO formulation for MLR minimizes over $\bB$ the worst-case expected loss: 
\begin{equation} \label{dro-MLR} 
\inf\limits_{\bB}\sup\limits_{\mbb{Q} \in \Omega}
\mbb{E}^{\mbb{Q}} \big[h_{\bB}(\bx, \by) \big],
\end{equation}
where \tcb{$\mbb{E}^{\mbb{Q}}$ denotes expectation under a  distribution $\mbb{Q}$ of the data $(\bx, \by)$, with $\mbb{Q}$ belonging to a set $\Omega$:}
\begin{equation} \label{Omega}
\Omega \triangleq \{\mbb{Q}\in \scrP(\scrZ):\ W_1(\mathbb{Q},\ \hat{\mathbb{P}}_N) \le \epsilon\},
\end{equation}
where $\scrZ$ is the set of possible values for $(\bx, \by)$, i.e., $\scrZ=\mbb{R}^p \times \{\mathbf{e}_1, \ldots, \mathbf{e}_K\}$; $\scrP(\scrZ)$ is
the space of all probability distributions supported on $\scrZ$; $\epsilon$ is a pre-specified positive constant; and $\hat{\mbb{P}}_N$ is the empirical distribution that assigns equal probability to each observed sample $(\bx_i, \by_i), i \in \lb N \rb$. 
$W_1(\mbb{Q},\ \hat{\mbb{P}}_N)$ is the order-1 Wasserstein distance
between $\mbb{Q}$ and $\hat{\mbb{P}}_N$:
\begin{equation} \label{wass1} 
W_1 (\mbb{Q}, \ \hat{\mbb{P}}_N) \triangleq \min\limits_{\Pi \in \scrP(\scrZ \times \scrZ)} \Bigl\{\int_{\scrZ \times \scrZ} l(\bz_1, \bz_2) \ \Pi \bigl(\mathrm{d}\bz_1, \mathrm{d}\bz_2\bigr)\Bigr\}, 
\end{equation}
where $\bz_i = (\bx_i, \by_i), i=1,2$, $\Pi$ is the joint distribution of $\bz_1$ and $\bz_2$ with
marginals $\mbb{Q}$ and $\hat{\mbb{P}}_N$, respectively, and $l(\cdot, \cdot)$ is a distance metric on the data space that measures the cost of transporting the probability mass and is defined as:
\begin{equation} \label{lg-dist}
l(\bz_1, \bz_2) = \|\bx_1 - \bx_2\|_r + M \|\by_1- \by_2\|_t,
\end{equation}
with a positive constant $M$. 

Theorem \ref{thm2} derives an equivalent reformulation of (\ref{dro-MLR}) by exploiting the dual problem of the inner supremum in (\ref{dro-MLR}). 
The proof can be found in the Supplement.

\begin{thm} \label{thm2}
	Suppose we observe $N$ realizations of the data, denoted by $(\bx_i, \by_i), i \in \lb N \rb$. When the Wasserstein metric is induced by (\ref{lg-dist}), as $M \rightarrow \infty$,
	the DRO problem (\ref{dro-MLR}) can be reformulated as:
	\begin{equation} \label{lg-relax1}
		\inf_{\bB} \frac{1}{N} \sum_{i=1}^N h_{\bB}(\bx_i, \by_i) + \epsilon 2^{1/s}  \|\bB\|_{s}, 
	\end{equation}
	where $r, s \ge 1$, and $1/r+1/s=1$. We call (\ref{lg-relax1}) the DRO-MLR formulation.
\end{thm}

\textbf{Remark: a weighted norm space.} When the feature space is equipped with a weighted norm, e.g.,
\begin{equation} \label{weight-lg-dist}
l(\bz_1, \bz_2) = \|\bx_1 - \bx_2\|_r^{\bW} + M \|\by_1- \by_2\|_t,
\end{equation}
where $\|\bx\|_r^{\bW} \triangleq \|\bW^{1/2}\bx\|_r$, with $\bW$ a positive definite matrix, the corresponding DRO-MLR formulation can be written as:
\begin{equation} \label{weight-dro-mlr}
		\inf_{\bB} \frac{1}{N} \sum_{i=1}^N h_{\bB}(\bx_i, \by_i) +  \epsilon 2^{1/s}  \|\bW^{-1/2}\bB\|_{s}, 
\end{equation}
where $r, s \ge 1$, $1/r+1/s=1$. This is due to the fact that the dual norm of $\|\cdot\|_r^{\bW}$ is simply $\|\cdot\|_s^{\bW^{-1}}$.

\section{Out-of-Sample Performance} \label{s3}
    In this section we \tcb{establish out-of-sample performance guarantees for the DRO-MLR solution, i.e., given a new test sample, we bound the expected log-loss of our prediction}. The resulting bounds shed light on the role of the regularizer in inducing a low prediction error. 
    


We want to measure the out-of-sample performance in terms of the empirical loss, which is typically used in {\em Empirical Risk Minimization (ERM)}, so that we can illustrate the advantage of DRO-MLR compared to ERM. By bounding the {\em Rademacher complexity} of the class of loss functions, Theorem \ref{c1} bounds the expected log-loss by the empirical loss plus additional terms that are inversely proportional to $\sqrt{N}$.

\begin{ass} \label{b1} For any $\bx$: \tcb{$\|\bx\|_s \le R \ \text{almost surely}$, for some scalar $R$.}
\end{ass}

\begin{ass} \label{b2} For any feasible solution $\bB$ to (\ref{lg-relax1}):
	$ \|\bB'\|_s \le \bar{C}$, \tcb{for some scalar $\bar{C}$.}
\end{ass}

With standardized predictors, \tcb{$R$} in Assumption \ref{b1} can be assumed to be small. The form of the constraint in Assumptions \ref{b2} is consistent with the form of the regularizers in DRO-MLR. We will see later that the bound $\bar{C}$ controls the out-of-sample log-loss of the solution to DRO-MLR, which validates the role of the regularizer in improving the out-of-sample performance. 

\begin{thm} \label{c1} Suppose the solution to the DRO-MLR formulation (\ref{lg-relax1}) is $\hat{\bB}_N$. Under Assumptions \ref{b1} and \ref{b2}, for any $0<\alpha<1$, with probability at least $1-\alpha$ with respect to the sampling, and with $\mbb{P}^*$ the true measure, 
	\begin{equation*} \small
		\begin{aligned}
		\mathbb{E}^{\mbb{P}^*}[h_{\hat{\bB}_N}(\bx,\by)]  \le  & \ \mathbb{E}^{\hat{\mathbb{P}}_N}[h_{\hat{\bB}_N}(\bx,\by)]  +\frac{2 \big(\log K + \bar{C} R (1 + K^{1/r})\big)}{\sqrt{N}} \\
		& +  
		\big(\log K + \bar{C} R (1 + K^{1/r})\big)\sqrt{\frac{8\log(\frac{2}{\alpha})}{N}}\ .
		\end{aligned}
	\end{equation*}
\end{thm}
Theorem \ref{c1} says that
the out-of-sample generalization (test) error of the DRO-MLR solution is bounded by the average training error
plus a bias term of order $1/\sqrt{N}$.
The bound in Theorem \ref{c1} demonstrates the validity of DRO-MLR in leading to a good out-of-sample performance. 

\section{DRO-MLR in Deep Image Classifiers} \label{s4}

We apply DRO-MLR to deep ViT based image classifiers,
and provide an efficient mechanism for inducing robustness to random and adversarial attacks in deep neural networks through applying DRO to a random layer at each epoch, which is computationally efficient and can be generalized to any deep learning-based approaches. We adopt a metric learning approach to estimate an appropriate norm for defining the Wasserstein metric. Our method is compared with ERM and other adversarial training methods 
to demonstrate the effectiveness of our model in inducing a smaller generalization error and test error rate.


     


\begin{figure*}
     \centering
     
     \begin{subfigure}[b]{\textwidth}
         \centering
         \includegraphics[width=\textwidth]{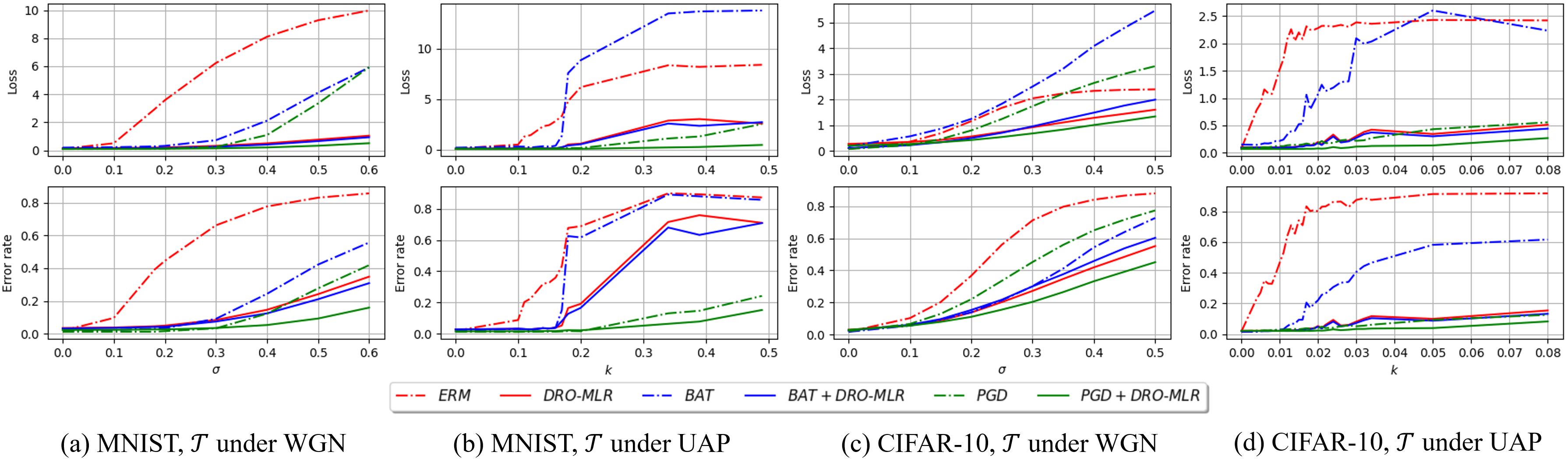}
         \label{fig:a}
     \end{subfigure}
    \vspace{-1em}
    \caption{Out-of-sample classification error and log-loss of different methods using ViT.}
    \label{fig2}
\end{figure*}

We split the dataset into a training set $\scrD$, a validation set $\scrV$ where hyper-parameter tuning (e.g., the regularization coefficient $\epsilon$) is performed, and a test set $\scrT$. A perturbed image $\tilde{\bx}$ is generated as $\tilde{\bx}=\bx+\bdelta$, where $\bdelta$ is generated by either a random attack such as {\em White Gaussian Noise (WGN)}, or an adversarial attack such as {\em Universal Adversarial Perturbation (UAP)}~\cite{UAP}. The WGN attack perturbs each pixel by a Gaussian noise with zero mean and standard deviation $\sigma$, while the UAP attack is generated using $10,000$ images from the training set $\scrD$, with $\|\bdelta\|_{\infty} \leq k$, and $k$ a pre-specified perturbation parameter. UAP has been shown to misclassify new images with high probability irrespective of the network architecture. 


We consider three baseline methods: (1) {\em Empirical Risk Minimization (ERM)}, which simply minimizes the sample averaged log-loss; (2) {\em Brute-force Adversarial Training (BAT)}, which adds adversarial samples into the training set $\scrD$ and performs ERM; and (3) {\em Projected Gradient Descent (PGD)} \cite{PGD}, which iteratively finds an optimal perturbation vector by using the gradient of the trained model from the previous iteration, and then updates the model using the perturbed samples. BAT is widely used in practice since it can be easily implemented, and PGD is known to be a strong defensive approach against many types of adversarial attacks. We compare these methods with their variations where DRO-MLR is applied, in terms of the log-loss and classification error on the test set $\scrT$.

We use $r=2$ to define the distance metric (\ref{lg-dist}). It is worth noting that when the DRO-MLR model is applied to a layer $l$, whose input is denoted as $\bphi_l(\bx)$ which takes into account the non-linear transformation of the raw image $\bx$ resulted from all layers before $l$, we need to estimate a proper distance metric in the transformed space to account for the distributional shift resulted from $\bphi_l$. We adopt a weighted norm metric as defined in (\ref{weight-lg-dist}) to approximate the effect of $\bphi_l$, and estimate the weight matrix $\bW$ by solving the following metric learning problem on the training set $\scrD$:
\begin{equation} \label{metric_learning}
	\begin{array}{rl}    
	\min\limits_{\bW\succcurlyeq 0} & \ \sum\limits_{\bx_i\in \scrD} \|\bW^{1/2}(\bphi_l (\tilde{\bx}_i)-\bphi_l(\bx_i))\|_2^2 \\
	\text{s.t.}  & \ \frac{1}{|\scrS|}\sum\limits_{(i,j)\in \scrS} \|\bW^{1/2}(\bphi_l (\tilde{\bx}_i)-\bphi_l(\tilde{\bx}_j))\|_2^2\geq c, \\
		 & \ \frac{1}{|\scrS|}\sum\limits_{(i,j)\in \scrS} \|\bW^{1/2}(\bphi_l (\bx_i)-\bphi_l(\bx_j))\|_2^2\geq c, \\
	\end{array}
\end{equation}
where $\tilde{\bx}_i$ is the perturbed version of $\bx_i$,  $\scrS \triangleq \{(i,j)|\bx_i, \bx_j \in \scrD, \by_i\neq \by_j\}$, $|\scrS|$ denotes the cardinality of the set $\scrS$, and $c$ is a fixed parameter. Note that for ViT where there exist multiple patches, we simply add up the squared norm in the objective and constraints of (\ref{metric_learning}) over patches.
Problem (\ref{metric_learning}) enforces that the distances between similar samples (evaluated in the transformed space $\bphi_l(\bx)$) are being minimized while the distances between dissimilar samples are sufficiently far away. (\ref{metric_learning}) is a semidefinite programming problem (SDP) which can be solved using SDPT3 solver~\cite{toh1999sdpt3}. In order to speed up the training, in each epoch, we only apply DRO-MLR to one random layer 
while keeping all other layers frozen, which has a similar flavor to the fine-tuning strategy of~\cite{girshick2014rich} that modifies the parameters of an existing network to train for a new task while preserving representations learned from the original task. Our novel random-layer training approach speeds up the training and bears a similar spirit to random dropout where efficiency and robustness are balanced through randomness.

\begin{table}[] \centering
\resizebox{\columnwidth}{!}{\begin{tabular}{@{}ccccccrrrr@{}}
\toprule
\multicolumn{1}{l}{} & \multicolumn{1}{l}{} & \multicolumn{4}{c}{MNIST}                                                                                 & \multicolumn{4}{c}{CIFAR-10}                                                                              \\ \midrule
\multicolumn{1}{l}{} & \multicolumn{1}{l}{} & \multicolumn{2}{c}{WGN}                             & \multicolumn{2}{c}{UAP}                             & \multicolumn{2}{c}{WGN}                             & \multicolumn{2}{c}{UAP}                             \\ \midrule
\multicolumn{1}{l}{} & \multicolumn{1}{l}{} & \multicolumn{1}{l}{Mean} & \multicolumn{1}{l}{Std.} & \multicolumn{1}{l}{Mean} & \multicolumn{1}{l}{Std.} & \multicolumn{1}{l}{Mean} & \multicolumn{1}{l}{Std.} & \multicolumn{1}{l}{Mean} & \multicolumn{1}{l}{Std.} \\ \midrule
\multirow{2}{*}{ERM} & Loss                 & 89.4\%                   & 0.1\%                    & 69.5\%                   & 1.5\%                    & 32.2\%                   & 0.8\%                    & 79.1\%                   & 0.2\%                    \\ \cmidrule(l){2-10} 
                     & Error rate           & 58.1\%                   & 1.3\%                    & 19.3\%                   & 0.8\%                    & 37.7\%                   & 0.5\%                    & 83.5\%                   & 0.3\%                    \\ \midrule
\multirow{2}{*}{BAT} & Loss                 & 83.6\%                   & 0.7\%                    & 79.5\%                   & 0.6\%                    & 64.4\%                   & 0.9\%                    & 78.3\%                   & 1.7\%                    \\ \cmidrule(l){2-10} 
                     & Error rate           & 42.8\%                   & 2.5\%                    & 19.8\%                   & 2.5\%                    & 15.8\%                   & 1.3\%                    & 76.3\%                   & 1.9\%                    \\ \midrule
\multirow{2}{*}{PGD} & Loss                 & 91.3\%                   & 0.2\%                    & 79.9\%                   & 1.5\%                    & 59.9\%                   & 0.9\%                    & 53.4\%                   & 1.3\%                    \\ \cmidrule(l){2-10} 
                     & Error rate           & 61.0\%                   & 2.3\%                    & 28.8\%                   & 7.3\%                    & 43.6\%                   & 1.7\%                    & 35.7\%                   & 1.5\%                    \\ \bottomrule
\end{tabular}
}

\smallskip
\caption{Percentage improvement (Mean and Standard deviation) of DRO-MLR over three baseline methods using ViT at maximum attack strength under WGN and UAP attacks on the test set. Percentage improvement is defined as the ratio of the change in the metric (error rate or loss) over the base metric value.}
\label{table1}
\end{table}

We demonstrate the results on MNIST \cite{mnist} and CIFAR-10 \cite{cf10}, which contain 50k/10k/10k and 40k/10k/10k training/validation/test samples, respectively. We apply a ViT with 4 attention layers on MNIST, and on CIFAR-10, we fine-tuned a ViT-B/16 model \cite{vit}. We used the learning rate of $1 \times 10^{-3}$ for MNIST and $1 \times 10^{-4}$  for CIFAR-10, while no weight decay was applied.
The results are shown in Fig. \ref{fig2}, where we plot the average log-loss $h_{\scrT}$ and classification error $e_{\scrT}$ on the test set $\scrT$ for various methods under different attacks, as the perturbation strength $\sigma$ or $k$ varies. 
We see that when DRO-MLR is combined with various adversarial training methods, both the loss and error rate are significantly reduced, with PGD+DRO-MLR outperforming the rest.
The performance gap becomes more prominent as the perturbation strength increases. 
The results suggest that DRO-MLR can be potentially combined with any existing adversarial training method on any neural network structure to further improve its performance. 

Table \ref{table1} summarizes the reduction in the loss and error rate at maximum $\sigma$ and $k$. On MNIST, when ERM / BAT is combined with DRO-MLR, the test error rate is reduced by up to 58\% / 43\% and log-loss is reduced by up to 89\% / 84\%, respectively. On CIFAR-10, the reductions w.r.t. ERM / BAT are 84\% / 76\% for error rate, and 79\% / 78\% for log-loss. Note that PGD remains a powerful defense that it can sometimes outperform the vanilla DRO-MLR model under the adversarial attack. Nevertheless, when combined with DRO-MLR, PGD can be further improved by up to 61\% / 91\% for error rate / log-loss on MNIST, and up to 44\% / 60\% on CIFAR-10. On the other hand, Fig. \ref{fig2} indicates that DRO-MLR has a comparable performance to ERM under very small perturbations, implying that DRO-MLR is able to induce robustness to perturbations without compromising the accuracy on unperturbed data. 

We want to emphasize that during the training, both DRO-MLR and BAT make use of the adversarial samples. The fact that DRO-MLR outperforms BAT suggests that DRO-MLR uses a more efficient way to leverage the information in the adversarial samples (to learn the $\bW$ matrix) without expanding the training set, which could also explain the significant improvement that DRO-MLR brings to PGD,
demonstrating the potential of DRO-MLR in being combined with other adversarial training methods to make further improvement. 

\begin{figure}
     \centering
     
     \begin{subfigure}[b]{0.235\textwidth}
         \centering
         \includegraphics[width=\textwidth]{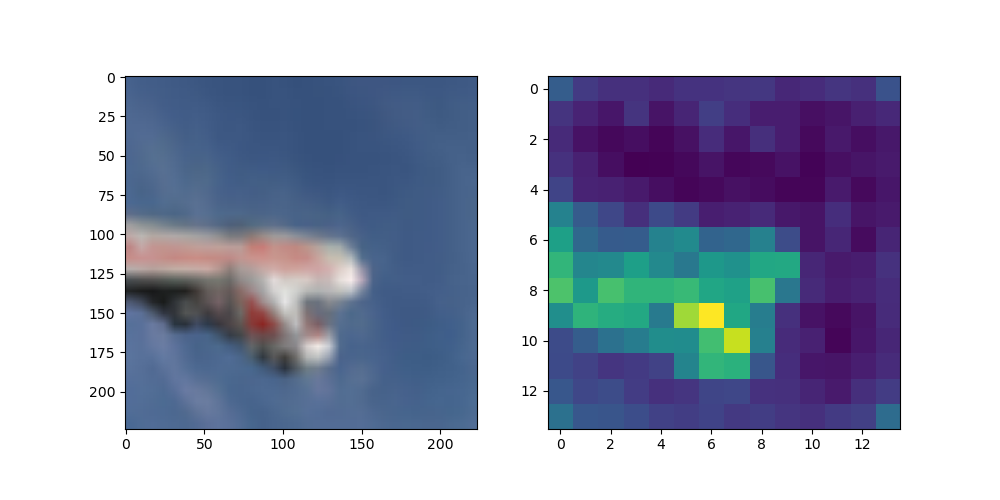}
         \vspace{-2em}
         \caption{ERM on clean image}
         \label{fig:a}
     \end{subfigure}
     \hfill
     \begin{subfigure}[b]{0.235\textwidth}
         \centering
         \includegraphics[width=\textwidth]{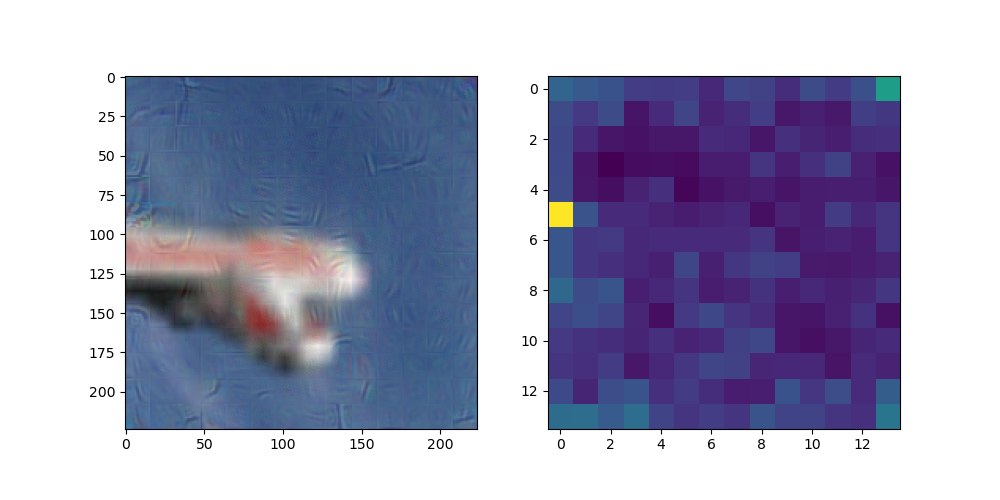}
         \vspace{-2em}
         \caption{ERM on UAP-attacked image}
         \label{fig:b}
     \end{subfigure}
     \hfill
     
     \begin{subfigure}[b]{0.235\textwidth}
         \centering
         \includegraphics[width=\textwidth]{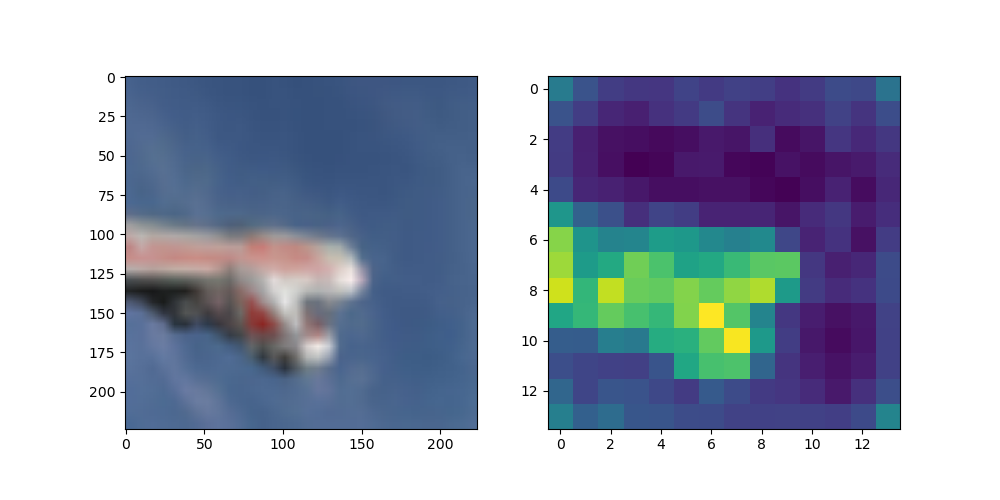}
         \vspace{-2em}
         \caption{DRO-MLR on clean image\\~}
         \label{fig:c}
     \end{subfigure}
     \hfill
     \begin{subfigure}[b]{0.235\textwidth}
         \centering
         \includegraphics[width=\textwidth]{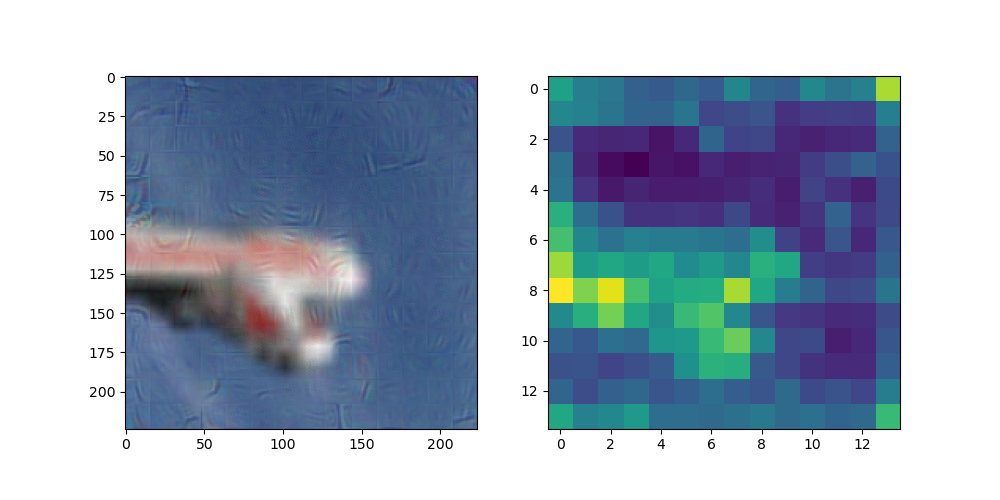}
         \vspace{-2em}
         \caption{DRO-MLR on UAP-attacked image}
         \label{fig:d}
     \end{subfigure}

        \caption{Visualization of the attention maps from the ViT.}
        \label{fig_attmap}
\end{figure}

Finally, visualizing the attention maps in ViT can also help to understand the advantages of DRO-MLR. In Fig. \ref{fig_attmap} (a) and (c), both the ERM and DRO-MLR models can capture the most important patches in the clean images through the self-attention mechanism. Nevertheless, when the image is slightly perturbed by the UAP attack in Fig. \ref{fig_attmap} (b), self-attention under ERM loses its strength and fails to capture the important area, while DRO-MLR can still clearly find the position of the airplane in Fig. \ref{fig_attmap} (d). This serves as a strong evidence that DRO-MLR can help ViT preserve reliable self-attention. 

\section{Conclusion} \label{s5}
We proposed a novel distributionally robust framework under the Wasserstein metric for {\em Multiclass Logistic Regression (MLR)},
and reformulated the min-max formulation to a regularized empirical
loss minimization problem, establishing a connection between robustness and regularization in the multivariate setting.
We provide both theoretical results on the performance of our estimator, and empirical evidence on deep ViT-based image classifiers, showing that our DRO-MLR model reduces the baseline test error by up to 83.5\% and loss by up to 91.3\% under random and adversarial attacks.

\newpage
\bibliographystyle{IEEEbib}
\bibliography{ref_mlr}

\section{Supplementary Material}


\subsection{Omitted Corollaries}

The following results are needed to establish Theorem 2.1 in the main paper.

\begin{col} \label{sup-conj}
	Define the convex log-loss in class $k$ as $h_{\bB}(\bx, \mathbf{e}_k) = \log \mathbf{1}'e^{\bB'\bx} - (\bB\mathbf{e}_k)'\bx$. We have:
	\begin{equation*}
	\sup_{\bx \in \mbb{R}^p} h_{\bB}(\bx,\mathbf{e}_k) - \lambda \|\bx - \bx_i\|_r =
	\begin{cases} 
	h_{\bB}(\bx_i, \mathbf{e}_k), & \text{if \ $\lambda \geq \kappa,$} \\
	+\infty, & \text{otherwise},
	\end{cases}
	\end{equation*}
	where $\kappa \triangleq \sup\{\|\bB (\bgamma -\mathbf{e}_k)\|_s: \bgamma \geq \mathbf{0}, \ \mathbf{1}'\bgamma = 1\}$, with $\mathbf{e}_k$ the $k$-th unit vector, and $r, s \geq 1$, $1/r+1/s=1$.
\end{col}
\begin{proof}
	The proof of Corollary \ref{sup-conj} uses the following result, which comes from the proof of Theorem 6.3 in \cite{esfahani2018data}.
	\begin{col} [\cite{esfahani2018data}, Theorem 6.3]  \label{convex-conj}
		Suppose the loss function $h(\bx)$ is convex in $\bx \in \mbb{R}^p$. We have:
		\begin{equation*}
		\sup_{\bx \in \mbb{R}^p} h(\bx) - \lambda \|\bx - \bx_i\|_r =
		\begin{cases} 
		h(\bx_i), & \text{if \ $\lambda \geq \kappa,$} \\
		+\infty, & \text{otherwise},
		\end{cases}
		\end{equation*}
		where $\kappa \triangleq \sup\{\|\btheta\|_s: h^*(\btheta)<\infty\}$, $r, s \geq 1$, $1/r+1/s=1$, and $h^*(\btheta)$ denotes the convex conjugate function of $h(\bx)$.
	\end{col}
	
	To prove Corollary \ref{sup-conj}, the key is to compute the value of $\kappa$. We define 
	$$h_{k}(\bx) \triangleq h_{\bB}(\bx, \mathbf{e}_k) = \log \mathbf{1}'e^{\bB'\bx} - \bw_k'\bx.$$
	The function $h_{k}(\bx)$ is convex in $\bx$, due to 
	the convexity of $\log \mathbf{1}'e^{\bB'\bx}$. From Corollary \ref{convex-conj} we see that, in order to compute $\kappa$, we need to find the convex conjugate of $h_{k}(\bx)$.
	To do so, we first compute the convex conjugate of $f(\bx) \triangleq \log \mathbf{1}'e^{\bB'\bx}$.
	\begin{equation} \label{conj}
	\begin{array}{rl}    
	f^*(\btheta) & \triangleq 
	\sup_{\bx \in \mbb{R}^p}\ \{\btheta'\bx-f(\bx)\} \\
	& = \sup_{\bx \in \mbb{R}^p}\ \{\btheta'\bx-\log \mathbf{1}'e^{\bB'\bx} \}. \\
	\end{array}
	\end{equation}
	Write the first-order condition of Problem (\ref{conj}) as:
	\begin{equation*}
	\btheta  - \frac{\bB e^{\bB'\bx^*}}{\mathbf{1}'e^{\bB'\bx^*}} = \mathbf{0},
	\end{equation*}
	where $\bx^*$ is the stationary point. This implies that
	\begin{equation*}
	\btheta = \bB \bgamma,  	
	\end{equation*}
	where $\bgamma = e^{\bB'\bx^*}/\mathbf{1}'e^{\bB'\bx^*}$. We thus have that:
	\begin{equation*}
	    f^*(\btheta) < \infty, \ \text{if \ $\btheta = \bB \bgamma, \ \text{where} \ \bgamma \geq \mathbf{0}, \ \mathbf{1}'\bgamma = 1$}. 
	\end{equation*}
	The convex conjugate of $h_{k}(\bx)$ can be expressed as:
	\begin{equation} \label{conj-3}
	\begin{aligned}
	h_{k}^*(\btheta) & \triangleq 
	\sup_{\bx \in \mbb{R}^p}\ \{\btheta'\bx-h_k(\bx)\} \\
	& = \sup_{\bx \in \mbb{R}^p}\ \{(\btheta+\bw_k)'\bx-\log \mathbf{1}'e^{\bB'\bx} \} \\
	& = f^*(\btheta+\bw_k). \\
	\end{aligned}
	\end{equation}
	To make $h_{k}^*(\btheta) < \infty$, it must satisfy that $\btheta+\bw_k =  \bB \bgamma$, where $\bgamma \geq \mathbf{0}, \ \mathbf{1}'\bgamma = 1$. Therefore, 
	\begin{equation*}
	    \begin{aligned}
	        \kappa & \triangleq \sup\{\|\btheta\|_s: h_k^*(\btheta)<\infty\} \\
	        & = \sup\{\|\bB \bgamma - \bw_k\|_s: \bgamma \geq \mathbf{0}, \ \mathbf{1}'\bgamma = 1\}.
	    \end{aligned}
	\end{equation*}
\end{proof}

\subsection{Omitted Proof to Theorem 2.1}

\begin{proof}
	Let us first examine the inner supremum of the DRO problem, which can be expressed as:
	\begin{equation} \label{inner-sup}
	   \sup\limits_{\mbb{Q} \in \Omega}
	\mbb{E}^{\mbb{Q}} [h_{\bB}(\bx, \by)]  = \sup\limits_{\mbb{Q} \in \Omega} \int\nolimits_{\scrZ} h_{\bB}(\bz) \mathrm{d} \mbb{Q}(\bz), 
	\end{equation}
	where $\bz = (\bx, \by)$. By definition of the Wasserstein set we can reformulate (\ref{inner-sup}) as:
	\begin{equation} \label{primal-1-rc}
	\begin{aligned}   
	\sup\limits_{\Pi \in \scrP(\scrZ \times \scrZ)} & \ \int\nolimits_{\scrZ} h_{\bB}(\bz) \mathrm{d} \Pi(\bz, \scrZ) \\
	\text{s.t.}  & \ \int\nolimits_{\scrZ \times \scrZ} l(\bz, \tcb{\tilde{\bz}})  \mathrm{d}\Pi(\bz, \tcb{\tilde{\bz}}) \le \epsilon, \\
	& \ \int\nolimits_{\scrZ \times \scrZ} \delta_{\bz_i}(\tcb{\tilde{\bz}})  \mathrm{d}\Pi(\bz,\tcb{\tilde{\bz}}) = \frac{1}{N}, \ \forall i \in \lb N \rb, \\
	\end{aligned}
	\end{equation}
	where $\Pi$ is the joint distribution of $\bz$ and $\tcb{\tilde{\bz}}$ with marginals $\mbb{Q}$ and $\hat{\mbb{P}}_N$, $\tcb{\tilde{\bz}}$ indexes the support of $\hat{\mbb{P}}_{N}$, and $\delta_{\bz_i}(\cdot)$ is the Dirac delta function at point $\bz_i$. Using $\mbb{Q}^i$ to denote the conditional distribution of $\bz$ given $\tcb{\tilde{\bz}}=\bz_i$, we can rewrite (\ref{primal-1-rc}) as:
	\begin{equation} \label{primal-2-rc}
	\begin{aligned}   
	\sup\limits_{\mbb{Q}^i} & \ \frac{1}{N} \sum\limits_{i=1}^N \int\nolimits_{\scrZ} h_{\bB}(\bz) \mathrm{d} \mbb{Q}^i(\bz) \\
	\text{s.t.}  & \ \frac{1}{N} \sum\limits_{i=1}^N \int\nolimits_{\scrZ} l(\bz, \tcb{\tilde{\bz}})  \mathrm{d}\mbb{Q}^i(\bz) \le \epsilon, \\
	& \ \int\nolimits_{\scrZ} \mathrm{d} \mbb{Q}^i(\bz) = 1, \ \forall i \in \lb N \rb. \\
	\end{aligned}
	\end{equation}
	Notice that the support $\scrZ$ can be decomposed into $\mbb{R}^p$ and a discrete set $\{\mathbf{e}_1, \ldots, \mathbf{e}_K\}$. We thus decompose each distribution $\mbb{Q}^i$ into unnormalized measures $\mbb{Q}^i_k$ supported on $\mbb{R}^p$ such that
	$\mbb{Q}^i_k (\mathrm{d}\bx) \triangleq \mbb{Q}^i(\mathrm{d}\bx, \by = \mathbf{e}_k), k \in \lb K \rb$. Problem (\ref{primal-2-rc}) can then be reformulated as:
	\begin{equation} \label{primal-3-rc}
	\begin{aligned}   
	\sup\limits_{\mbb{Q}^i_k} & \ \frac{1}{N} \sum\limits_{i=1}^N  \sum\limits_{k=1}^K \int\nolimits_{\mbb{R}^p} h_{\bB}(\bx, \mathbf{e}_k) \mathrm{d} \mbb{Q}^i_k(\bx) \\
	\text{s.t.}  & \ \frac{1}{N} \sum\limits_{i=1}^N  \sum\limits_{k=1}^K \int\nolimits_{\mbb{R}^p} l((\bx, \mathbf{e}_k), (\bx_i, \by_i))  \mathrm{d}\mbb{Q}^i_k(\bx) \le \epsilon, \\
	& \  \sum\limits_{k=1}^K \int\nolimits_{\mbb{R}^p} \mathrm{d} \mbb{Q}^i_k(\bx) = 1, \ \forall i \in \lb N \rb. \\
	\end{aligned}
	\end{equation}
	Using the definition of $l$, we can write Problem (\ref{primal-3-rc}) as:
		\begin{equation*} 
		\begin{aligned}
		\sup\limits_{\mbb{Q}^i_k} & \ \frac{1}{N} \sum\limits_{i=1}^N  \sum\limits_{k=1}^K \int\nolimits_{\mbb{R}^p} h_{\bB}(\bx, \mathbf{e}_k) \mathrm{d} \mbb{Q}^i_k(\bx) \\
		\text{s.t.}  & \ \frac{1}{N} \sum\limits_{i=1}^N  \sum\limits_{k=1}^K \int\nolimits_{\mbb{R}^p} \Big(\|\bx - \bx_i\|_r + M \|\mathbf{e}_k- \by_i\|_t \Big)  \mathrm{d}\mbb{Q}^i_k(\bx) \le \epsilon, \\
		& \  \sum\limits_{k=1}^K \int\nolimits_{\mbb{R}^p} \mathrm{d} \mbb{Q}^i_k(\bx) = 1, \ \forall i \in \lb N \rb, \\
		\end{aligned}
		\end{equation*}
		which can be equivalently written as:
		\begin{equation} \label{primal-4-rc}
		\begin{aligned}   
		\sup\limits_{\mbb{Q}^i_k} & \ \frac{1}{N} \sum\limits_{i=1}^N  \sum\limits_{k=1}^K \int\nolimits_{\mbb{R}^p} h_{\bB}(\bx, \mathbf{e}_k) \mathrm{d} \mbb{Q}^i_k(\bx) \\
		\text{s.t.}  & \ \frac{1}{N} \int\nolimits_{\mbb{R}^p} \Bigl(\sum\limits_{i=1}^N \|\bx - \bx_i\|_r \Big( \sum\limits_{k=1}^K \mathrm{d}\mbb{Q}^i_k(\bx)\Big)  \\
		& \ +  2^{1/t}M \Big(\sum\limits_{k} \sum\limits_{i: \by_i=\mathbf{e}_k} \mathrm{d}\overline{\mbb{Q}}^i_k(\bx) \Big)\Bigr)\le \epsilon, \\
		& \  \sum\limits_{k} \int\nolimits_{\mbb{R}^p} \mathrm{d} \mbb{Q}^i_k(\bx) = 1, \ \forall i \in \lb N \rb, \\
		\end{aligned}
		\end{equation}
		where $\overline{\mbb{Q}}^i_k \triangleq \sum_{j=1}^K \mbb{Q}^i_j - \mbb{Q}^i_k$. In the derivation we used the fact that $\|\mathbf{e}_i - \mathbf{e}_j\|_t = 2^{1/t}$, if $i\neq j$.
		
		Notice that (\ref{primal-4-rc}) is a linear problem (LP) in $\mbb{Q}_k^i$. We can apply linear duality with dual variables $\lambda$ and $s_i$. (\ref{primal-4-rc}) is a special LP in that the decision variables $\mbb{Q}_k^i$ are infinite dimensional. For each $\mbb{Q}_k^i(\bx)$, its coefficients in the constraints of (\ref{primal-4-rc}) are multiplied by the corresponding dual variables to produce the constraints of the dual problem (\ref{dual-1-rc}) (LP duality). Since $\mbb{Q}_k^i$ has infinitely many arguments $\bx$, the constraints of (\ref{dual-1-rc}) involve the supremum over $\bx$. The dual problem of (\ref{primal-4-rc}) can be written as:
		\begin{equation} \label{dual-1-rc}
		\begin{aligned}
			\inf\limits_{\substack{\lambda \geq 0, s_i}} & \ \lambda \epsilon + \frac{1}{N} \sum\limits_{i=1}^N s_i \\
			\text{s.t.} & \ \sup\limits_{\bx \in \mbb{R}^p} h_{\bB}(\bx, \mathbf{e}_k) - \lambda \|\bx - \bx_i\|_r - \lambda M \|\mathbf{e}_k - \by_i\|_t \leq s_i, \\ & \ \forall i \in \lb N \rb, k \in \lb K \rb.
		\end{aligned}
		\end{equation}
	Note that the value of Problem \ref{dual-1-rc} is equal to the value of \ref{primal-4-rc} for the optimal dual variable $\lambda$, due to strong duality. Using Corollary 1.1, we can write Problem (\ref{dual-1-rc}) as:
	\begin{equation} \label{dual-2-rc}
	\begin{aligned}
	\inf\limits_{\lambda, s_i} & \ \lambda \epsilon + \frac{1}{N} \sum\limits_{i=1}^N s_i \\
	\text{s.t.} & \ h_k(\bx_i) - \lambda M \|\mathbf{e}_k - \by_i\|_t \leq s_i, \ \forall i \in \lb N \rb, k \in \lb K \rb, \\
	& \ \lambda \geq \sup\{\|\bB (\bgamma -\mathbf{e}_k)\|_s: \bgamma \geq \mathbf{0}, \ \mathbf{1}'\bgamma = 1\},\ 			\tcb{\forall k \in \lb K \rb},
	\end{aligned}
	\end{equation}
	where $1/r + 1/s=1$.
	As $M\rightarrow \infty$, i.e., we assign a very large
weight on the labels, implying that samples from different classes are
infinitely far away, the first set of constraints in Problem (\ref{dual-2-rc}) reduces to:
	$h_B(\bx_i, \by_i) \leq s_i, \ \forall i \in \lb N \rb.$
	Therefore, the optimal value of (\ref{dual-2-rc}) is:
	\begin{equation}  \label{MLR-reg}
	 \frac{1}{N} \sum\limits_{i=1}^N h_B(\bx_i, \by_i) + \lambda \epsilon,
	\end{equation}
	where $\lambda = \tcb{\max_k} \sup\{\|\bB (\bgamma - \mathbf{e}_k)\|_s: \bgamma \geq \mathbf{0}, \ \mathbf{1}'\bgamma = 1\}$. Note that by setting $M\rightarrow \infty$, it does not imply that we only care about the perturbation on the labels; instead, when the samples are in the same class, we focus on perturbations on the input feature $\bx$. To compute $\lambda$, notice that
	\begin{equation*}
	\|\bB (\bgamma - \mathbf{e}_k)\|_s 
		 \leq \|\bB\|_s  \|\bgamma - \mathbf{e}_k\|_s,
	\end{equation*}
	where $\|\bB\|_s$ is the induced $\ell_s$ norm of the matrix $\bB$.
	The maximum of $\|\bgamma - \mathbf{e}_k\|_s$ can be obtained as:
	\begin{equation*}
	\begin{aligned}
		\|\bgamma - \mathbf{e}_k\|_s^s & = \sum\nolimits_{i=1}^{k-1} \gamma_i^s + (1-\gamma_k)^s + \sum\nolimits_{j=k+1}^K \gamma_{j}^s \\
		& \leq \sum\nolimits_{i=1}^{k-1}\gamma_i + (1-\gamma_k) + \sum\nolimits_{j=k+1}^K \gamma_{j} \\
	& = 1 - 2 \gamma_k + 1 \\
	& \leq 2,
	\end{aligned}
	\end{equation*}
    where $\bgamma = (\gamma_1, \ldots, \gamma_K)$. Therefore, \tcb{(\ref{MLR-reg}) can be reformulated into Eq. (5) in the statement of Theorem 2.1 in the main paper by setting} $\lambda = 2^{1/s}\|\bB\|_s$.
\end{proof}

\subsection{Omitted Proof to Theorem 3.2}

\begin{proof}
     We need to find an upper bound to the loss $h_{\bB}(\bx, \by) = \log \mathbf{1}'e^{\bB'\bx} - \by'\bB'\bx$. To do this, we first bound $|h_{\bB}(\bx, \by) - h_{\bB}(\bx_0, \by)|$ for any $\bx_0 \in \mbb{R}^p$. Note that 
    \begin{equation} \label{log-loss-diff}
    \begin{aligned}
      & |h_{\bB}(\bx, \by) - h_{\bB}(\bx_0, \by)| 
     \\= & \ |\log \mathbf{1}'e^{\bB'\bx} - \by'\bB'\bx - \log \mathbf{1}'e^{\bB'\bx_0} + \by'\bB'\bx_0| \\
    \le & \ |\log \mathbf{1}'e^{\bB'\bx} - \log \mathbf{1}'e^{\bB'\bx_0}| + |\by'\bB'\bx - \by'\bB'\bx_0|.
    \end{aligned}
    \end{equation}
Let us examine the two terms in (\ref{log-loss-diff}) separately. For the first term, define a function $g(\ba) = \log \mathbf{1}'e^{\ba}$, where $\ba \in \mbb{R}^K$. Using the mean value theorem, we know for any $\ba, \bb \in \mbb{R}^K$, there exists some $t \in (0, 1)$ such that
\begin{equation}\begin{aligned}
\label{mvt}
|g(\mathbf{b}) - g(\ba)| \le & \bigl\|\nabla g\bigl((1-t)\ba + t\mathbf{b}\bigr)\bigr\|_r \|\mathbf{b} - \ba\|_s \\
\le & K^{1/r} \|\mathbf{b} - \ba\|_s,
\end{aligned}
\end{equation}
where $r, s \ge 1$, $1/r+1/s=1$, the first inequality is due to H\"{o}lder's inequality, and the second inequality is due to the fact that $\nabla g(\ba) = e^{\ba}/\mathbf{1}'e^{\ba}$, which implies that each element of $\nabla g(\ba)$ is smaller than 1. Based on (\ref{mvt}) we have:
\begin{equation}\begin{aligned} \label{first-term}
& |\log \mathbf{1}'e^{\bB'\bx} - \log \mathbf{1}'e^{\bB'\bx_0}| \le K^{1/r} \|\bB'(\bx-\bx_0)\|_s \\
\le & K^{1/r} \|\bB'\|_s \|\bx-\bx_0\|_s,
\end{aligned}
\end{equation}
where $r, s \ge 1$, $1/r+1/s=1$, and the last inequality is due to the definition of the matrix norm. For the second term of (\ref{log-loss-diff}), we have,
\begin{equation} \label{second-term}
 |\by'\bB'\bx - \by'\bB'\bx_0| 
\le \|\by\|_r \|\bB'(\bx - \bx_0)\|_s 
\le \|\bB'\|_s \|\bx - \bx_0\|_s,
\end{equation}
where the first inequality is due to H\"{o}lder's inequality, and the second inequality is due to the definition of the matrix norm and the fact that $\|\by\|_r = 1$. Combining (\ref{first-term}) and (\ref{second-term}), we have: 
\begin{equation*} 
\begin{aligned}
&|h_{\bB}(\bx, \by) - h_{\bB}(\bx_0, \by)| 
    \\ \le & K^{1/r} \|\bB'\|_s \|\bx-\bx_0\|_s + \|\bB'\|_s \|\bx - \bx_0\|_s.
    \end{aligned}
\end{equation*}
Under Assumptions A and B, by setting $\bx_0 = \mathbf{0}$, we obtain that,
\begin{equation*} 
      |h_{\bB}(\bx, \by) - h_{\bB}(\mathbf{0}, \by)| 
    \le  K^{1/r} \bar{C} R + \bar{C} R.
\end{equation*}
By noting that $h_{\bB}(\mathbf{0}, \by) = \log K$, we conclude:
$h_{\bB}(\bx, \by) \le \log K + \bar{C} R (1 + K^{1/r})$.

With the above results, the idea is to bound the expected loss using the empirical {\em Rademacher complexity} $\scrR_N(\cdot)$ of
the class of loss functions: 
$\scrH=\{(\bx, \by) \ra h_{\bB}(\bx, \by)\}$, denoted by $\scrR_N(\scrH)$. Using Lemma 4.3.2 of \cite{OPT-026} and the upper bound on the loss function, we arrive at the following result.

\begin{lem} \label{radcom-mlr}
	Under Assumptions A and B,
	\begin{equation*}
	\scrR_N(\scrH)\le \frac{2 (\log K + \bar{C} R (1 + K^{1/r})) }{\sqrt{N}}.
	\end{equation*}
\end{lem}
Using the Rademacher complexity of the class of loss functions, the out-of-sample prediction bias in Theorem 3.2 can be bounded by applying Theorem 8 in \cite{Peter02}.
\end{proof}

\subsection{Experimental settings}

We run the experiments on local GPU workstations with 4 NVIDIA RTX A6000 (48GB VRAM) and 2 NVIDIA Titan RTX (24GB VRAM) GPUs. The experiment for one epoch of DRO-MLR training on MNIST took only a few seconds while on CIFAR-10 it took about 0.05 GPU hours. Our ViT models were constructed under Huggingface Transformers v4.5.1 \cite{huggingface}.

\subsection{Omitted Experimental Results} 

\begin{figure*}
     \centering
     
     \begin{subfigure}[b]{\textwidth}
         \centering
         \includegraphics[width=\textwidth]{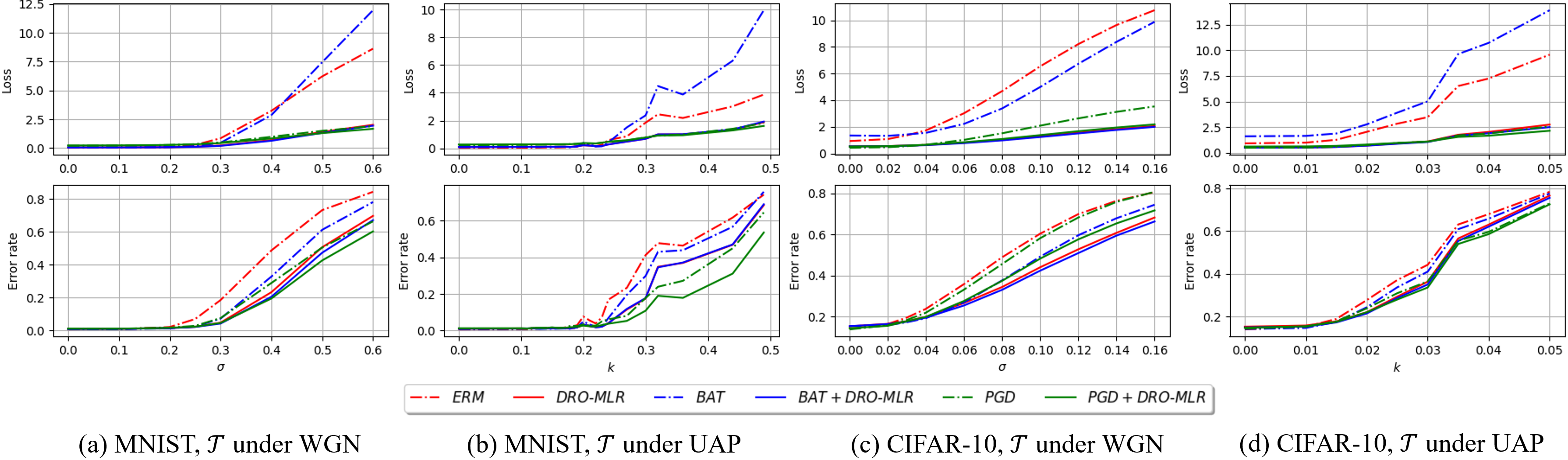}
         \label{fig_cnn:a}
     \end{subfigure}
    \vspace{-1em}
    \caption{Out-of-sample classification error and log-loss of different methods using CNN.}
    \label{fig_cnn}
\end{figure*}

We also implement DRO-MLR to Convolutional Neural Network (CNN) models. For a CNN image classifier, DRO-MLR is applied only to the last layer. We use a 10-layer Residual Network (ResNet) \cite{resnet} on MNIST, and a 18-layer ResNet on CIFAR-10. The performance improvement of DRO-MLR shown in Fig. \ref{fig_cnn} is less significant compared to that in the ViT models, due to the fact that we only apply DRO-MLR to the last layer of CNN, while for ViT, DRO-MLR is applied to a larger set of layers.

\begin{figure*}
     \centering
     \begin{subfigure}[b]{1\textwidth}
         \centering
         \includegraphics[width=\textwidth]{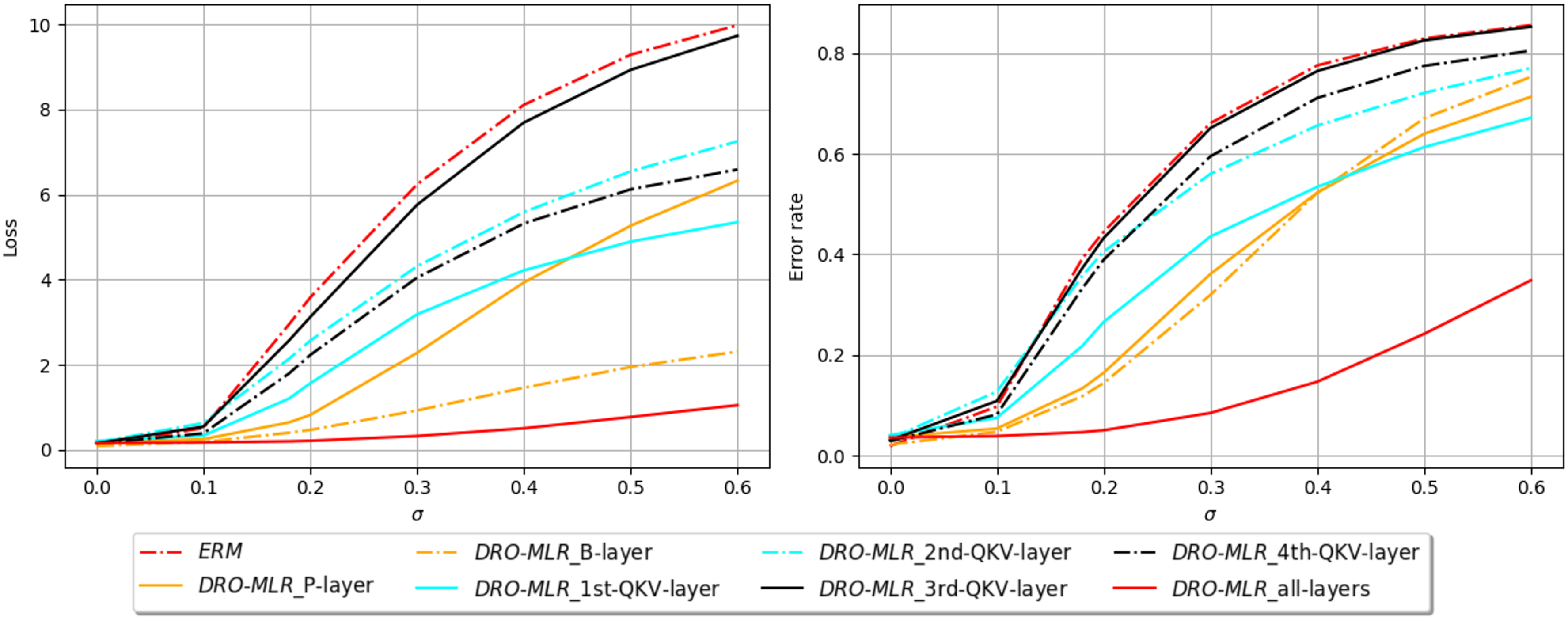}
         \label{fig_layers:a}
     \end{subfigure}
     \vspace{-3em}
    \caption{Performance of applying DRO-MLR to different layers of ViT on MNIST under WGN.}
    \label{fig_layers}
\end{figure*}

Finally, we briefly analyze the effect of applying DRO-MLR to different layers of ViT. In Fig. \ref{fig_layers}, DRO-MLR is applied separately to the final linear layer $B$, the initial patch projection layer $P$ or the $QKV$-mapping layer in one of the self-attention layers. Compared with ERM, not all layers bring a significant performance improvement. The overall performance boost when all layers are re-trained with DRO-MLR can be largely credited to the $B$ layer (the final linear layer). When DRO-MLR is applied only to the $B$ layer, the loss is reduced by up to 87.0\%, and the error rate is reduced by up to 67.6\%, showing that re-training only the last linear layer using DRO-MLR is a fast and reliable way to improve the robustness of existing methods (as we did in CNN).

\end{document}

%% file: mymacros_2cln.tex
\newcommand{\bemul}{\begin{multline*}}
\newcommand{\bee}{\begin{eqnarray*}}
\newcommand{\eee}{\end{eqnarray*}}
\newcommand{\been}[1]{\begin{eqnarray}\label{#1}}
\newcommand{\eeen}{\end{eqnarray}}
\newcommand{\began}[1]{\begin{gather}\label{#1}}
\newcommand{\bega}{\begin{gather*}}
\newcommand{\bealn}[1]{\begin{align}\label{#1}}
\newcommand{\beal}{\begin{align*}}
\newcommand{\bealatn}[2]{\begin{alignat}{#1}\label{#2}}
\newcommand{\bealat}{\begin{alignat*}}
\newcommand{\bexalatn}[1]{\begin{xalignat}\label{#1}}
\newcommand{\bexalat}{\begin{xalignat*}}

\newcommand{\ra}{\rightarrow}

\newcommand{\mbb}{\mathbb}

\newtheorem{thm}{Theorem}[section]

\newtheorem{lem}[thm]{Lemma}
\newtheorem{col}[thm]{Corollary}

\newtheorem{ass}{Assumption} 
 
\newcommand{\tc}{\textcolor}

\newcommand{\tcb}[1]{\tc{black}{#1}}

%% file: lpsymbols.tex
\def\ba{{\mathbf a}}
\def\bb{{\mathbf b}}

\def\bw{{\mathbf w}}
\def\bx{{\mathbf x}}  
\def\by{{\mathbf y}}
\def\bz{{\mathbf z}}
\def\bA{{\mathbf A}}
\def\bB{{\mathbf B}}

\def\bW{{\mathbf W}}

\def\texitem#1{\par\smallskip\noindent\hangindent 25pt
               \hbox to 25pt {\hss #1 ~}\ignorespaces}

\newcommand{\scrD}{\mathcal{D}}

\newcommand{\scrH}{\mathcal{H}}

\newcommand{\scrP}{\mathcal{P}}

\newcommand{\scrR}{\mathcal{R}}
\newcommand{\scrS}{\mathcal{S}}
\newcommand{\scrT}{\mathcal{T}}

\newcommand{\scrV}{\mathcal{V}}

\newcommand{\scrZ}{\mathcal{Z}}

\newcommand{\bgamma}{\boldsymbol{\gamma}}
\newcommand{\bdelta}{{\boldsymbol{\delta}}}

\newcommand{\btheta}{\boldsymbol{\theta}}

\newcommand{\bphi}{{\boldsymbol{\phi}}}